%% file: egpaper_for_review.tex
\documentclass[10pt,twocolumn,letterpaper]{article}

\usepackage{ijcb}
\usepackage{times}
\usepackage{epsfig}
\usepackage{graphicx}
\usepackage{amsmath}
\usepackage{amssymb}
\usepackage{multirow}
\usepackage{pifont}
\usepackage{hyperref}

\usepackage{subcaption}

\ijcbfinalcopy 

\ifijcbfinal\fi
\begin{document}

\title{Temporal Feature Alignment in Contrastive Self-Supervised Learning for Human Activity Recognition}

\author{Bulat Khaertdinov and Stylianos Asteriadis \\
Department of Advanced Computing Sciences, Maastricht University\\
Maastricht, Netherlands \\
{\tt\small \{b.khaertdinov, stelios.asteriadis\}@maastrichtuniversity.nl}
}

\maketitle
\thispagestyle{empty}

\begin{abstract}
   Automated Human Activity Recognition has long been a problem of great interest in human-centered and ubiquitous computing. In the last years, a plethora of supervised learning algorithms based on deep neural networks has been suggested to address this problem using various modalities. While every modality has its own limitations, there is one common challenge. Namely, supervised learning requires vast amounts of annotated data which is practically hard to collect. In this paper, we benefit from the self-supervised learning paradigm (SSL) that is typically used to learn deep feature representations from unlabeled data. Moreover, we upgrade a contrastive SSL framework, namely SimCLR, widely used in various applications by introducing a temporal feature alignment procedure for Human Activity Recognition. Specifically, we propose integrating a dynamic time warping (DTW) algorithm in a latent space to force features to be aligned in a temporal dimension. Extensive experiments have been conducted for the unimodal scenario with inertial modality as well as in multimodal settings using inertial and skeleton data. According to the obtained results, the proposed approach has a great potential in learning robust feature representations compared to the recent SSL baselines, and clearly outperforms supervised models in semi-supervised learning. The code for the unimodal case is available via the following link: \url{https://github.com/bulatkh/csshar_tfa}.
\end{abstract}

\input{sections/1_introduction}
\input{sections/2_related_work}
\input{sections/3_methodology}
\input{sections/4_implementation_details}
\input{sections/5_evaluations}
\input{sections/6_conclusion}
\input{sections/7_acknowledgement}

{\small
\bibliographystyle{ieee}
\bibliography{egbib}
}

\end{document}

%% file: sections/1_introduction.tex
\section{Introduction}
Movements and activities of humans provide crucial information that can be used to understand their habits and motives, monitor physical and mental health, analyze their performance in specific tasks as well as assist their daily life. The task of recognizing activities of people using data describing their movement is generally known as Human Activity Recognition (HAR). HAR is a problem that has been extensively addressed in many areas, such as smart homes \cite{Skocir_2016}, health monitoring \cite{Panwar_2019} and manufacturing automation \cite{GUNTHER_20191177}. 

Human Activity Recognition could be tackled using different sources of input data, such as wearable devices, RGB-D streams, skeletal joints and others. While methods based on individual modalities have their own weaknesses that can be neglected by the multimodal approaches, there is a significant practical challenge that has to be addressed. Namely, data annotation is an expensive and time-consuming process, especially for multimodal approaches. What is more, the recent supervised models are trained on large amounts of annotated data. 

Recent advances in representation learning have given rise to a new family of techniques, namely, Self-Supervised Learning (SSL). Through an SSL strategy, the goal is to build data representations in the absence of annotated samples by solving an auxiliary pre-text task. Such representations are subsequently utilized as inputs in order to train small-scale classifiers during the fine-tuning phase, necessitating only limited amounts of annotated data.

In the last years, contrastive SSL frameworks have drawn attention of researchers by demonstrating the state-of-the-art performance in various application areas \cite{Chen_2020_simclr, Chen_2021_simsiam}. The main idea of contrastive learning is to train a feature encoder to group semantically similar, or positive, data points together in a latent space, and push away the negative ones. When data annotations are not available, contrastive learning uses two different representations of the same input instance as a positive pair. Different representations of data could be obtained using augmentations in the unimodal case, while in the multimodal case these representations could be derived from different modalities \cite{Tian_2020_cmc}.

In this paper, inspired by \cite{Haresh2021LearningBA}, we aim to upgrade the contrastive learning frameworks and make use of the temporal nature of HAR task by introducing a Temporal Feature Alignment (TFA) algorithm that can be integrated into these frameworks. Specifically, the contributions of this paper could be summarized as follows: 
\begin{itemize}
    \item We propose to integrate a differentiable version of the Dynamic Time Warping (DTW) algorithm into contrastive learning frameworks applied to Human Activity Recognition to force alignment of features along the temporal dimension. 
    \item The proposed method is applicable on both unimodal and multimodal contrastive learning problems. In particular, we integrated them into the SimCLR \cite{Chen_2020_simclr} and Contrastive Multiview Coding (CMC) \cite{Tian_2020_cmc} frameworks.
    \vspace{-4pt}
    \item Extensive experiments have been conducted on unimodal sensor-based and multimodal (inertial and skeleton) HAR datasets. The obtained results have shown that the proposed method improves feature representation learning comparing with the recent SSL baselines, and, most importantly, with SimCLR and CMC, in multiple evaluation scenarios. 
\end{itemize}

%% file: sections/2_related_work.tex
\section{Related Work}

\subsection{Unimodal and Multimodal HAR}
Most of the algorithms applied on sensorial data obtained using IMUs address HAR as a multivariate time-series classification task. Hence, the applied deep learning methods used for the problem include such architectures as 1D-CNNs \cite{Yang_2015}, RNNs \cite{Hammerla_2016, Zhao_2018} and their combinations \cite{Ordonez_2016}. Moreover, attention mechanisms have been exploited in various forms, such as sensor attention, temporal attention and transformer self-attention \cite{Zeng_2018, Mahmud_2020, khaertdinov_2021}. Feature encoders for skeleton modality are typically based on either 2D-CNNs, RNNs or Graph Neural Networks \cite{li_2018_cooccurrence, song_2017_stlstm, yan_2018_stgcn, cheng_2020_shiftgcn}. In this paper we use a transformer-like architecture to encode inertial data and an adaptation of a so-called co-occurence feature learning architecture from \cite{li_2018_cooccurrence} for skeleton modality.

Mutlimodal HAR approaches apply modality fusion on varying levels, such on the raw data, on unimodal weak decisions, or on feature representations \cite{khaire_2018_combiningcnn, memmesheimer_2020_gimme_signals, das_2021_mmhar_ensemnet}. Other recent works propose more sophisticated end-to-end architectures that are crafted specifically for multimodal HAR. These approaches make use of recent advances in deep learning, such as GANs to generate feature representations of one modality given features from another \cite{wang_2019_generative_har}, various attention-based mechanisms to fuse different modalities together \cite{islam_2020_hamlet, islam_2021_multigat}, or knowledge distillation techniques \cite{liu_2021_sakdn}. In this paper, we make use of simple feature level fusion during fine-tuning in order to fairly assess the quality of feature representations learnt by individual encoders in a SSL manner.

\subsection{Contrastive Self-supervised Learning}
In the recent years, contrastive learning methods have shown the impressive performance in various applications, including unimodal activity recognition \cite{haresamudram_2021_cpc_har, Khaertdinov_2021_csshar}, by narrowing the gap between supervised and self-supervised methods. The main idea of this family of SSL methods is similar to metric learning, namely encoders are trained to group semantically similar, positive, examples. In case when no labels are available, positive examples are formed by crafting two different views from each instance. Moreover, some of the approaches use negative pairs that are used to avoid trivial collapsing solutions \cite{Tian_2020_cmc, Chen_2020_simclr, he_2020_moco}, while others propose not using them by introducing various schemes to prevent the trivial shortcuts \cite{grill_2020_byol, Chen_2021_simsiam}. A recent study on video understanding by Haresh and Kumar et al. \cite{Haresh2021LearningBA} proposed to align semantically similar video frames in time using the soft Dynamic Time Warping algorithm \cite{cuturi2017softdtw} and additional regularization that prevents their encoders from collapsing. Inspired by this idea, in this paper, we propose to exploit the nature of data used for HAR and attempt aligning features along temporal dimension by integrating a soft version of DTW into contrastive learning frameworks used in unimodal and multimodal settings, namely SimCLR \cite{Chen_2020_simclr} and CMC \cite{Tian_2020_cmc}.

%% file: sections/3_methodology.tex
\section{Methodology}
\begin{figure*}[!t]
\centering
\begin{minipage}{0.9\linewidth}
\centering
\subfloat[Unimodal contrastive learning based on inertial data.]{\label{fig:csshar}\includegraphics[scale=.3]{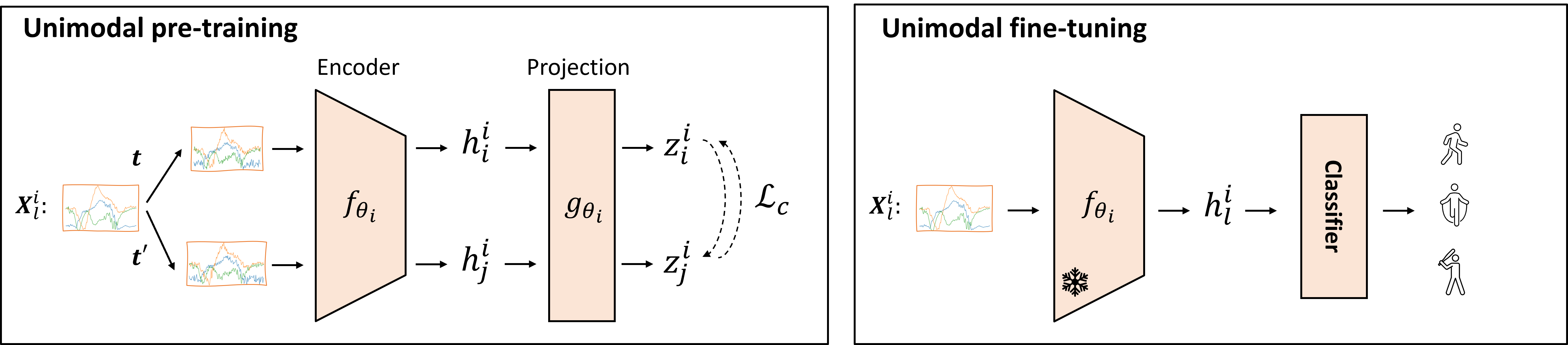}}
\end{minipage}%
\hfill
\centering
\begin{minipage}{0.9\linewidth}
\centering
\subfloat[Contrastive Multiview Coding (CMC) adapted to multimodal HAR using inertial and skeleton data.]{\label{fig:cmc}\includegraphics[scale=.3]{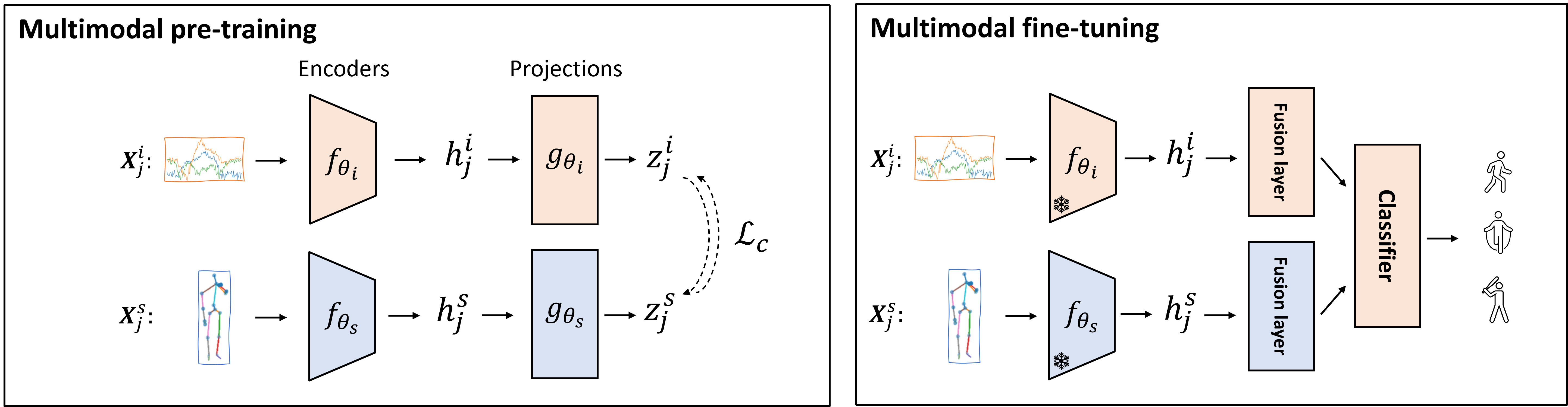}}
\end{minipage}
\caption{Unimodal and multimodal contrastive learning approaches adapted to inertial and inertial-skeleton HAR.}
\vspace{-10pt}
\end{figure*}

\subsection{Problem Definition}
Inertial signals, used in sensor-based HAR, are normally obtained from wearable sensors such as accelerometers, gyroscopes and others. Thus, sensor-based HAR can be considered a multivariate time-series classification task. Specifically, at timestamp $t$, the input signal is defined as $\boldsymbol{x_t} = [x_t^1, x_t^2, \ldots, x_t^S] \in \mathbb{R}^{S}$, where $S$ is a number of channels. Hence, a time-window with signals aggregated for $T$ timesteps can be written as $\boldsymbol{X}^i = [\boldsymbol{x_1}, \boldsymbol{x_2}, \ldots, \boldsymbol{x_T}]$. Finally, the goal is to associate each time-window with a correct output label $y \in Y$. 

Similarly, in this paper we address multimodal HAR using inertial sensors $\boldsymbol{X}^i$ and 2D skeletal joints $\boldsymbol{X}^s$. Skeleton data is generally represented as a set of coordinates tracked over time for a number of keypoints on a body. For any skeleton sequence, we denote $T$ as the number of frames in the sequence, $J$ as the number of joints and $C = 2$ as the number of data channels, or dimensionality of coordinates. Then, a skeleton sequence $\boldsymbol{X}^s \in \mathbb{R}^{T \times J \times C}$ consists of $T$ frames where the skeleton data for each frame is described by $\boldsymbol{P}_t = [\boldsymbol{p}_t^1, \boldsymbol{p}_t^2, \ldots, \boldsymbol{p}_t^J]$ and $\boldsymbol{p}_t^j \in \mathbb{R}^C$ is the position of joint $j$ at frame $t$.

\subsection{Contrastive SSL for HAR}
The vast majority of self-supervised learning frameworks are divided into two stages, namely pre-training and fine-tuning. The aim of pre-training, also referred to as a pre-text task, is to train a feature encoder on an auxiliary task derived from unlabeled data. In contrastive learning pre-training, the aim is to group semantically similar inputs in a feature space and split different examples apart. Since in SSL data annotations are not available, semantic similarity is defined by instance information. Specifically, the goal is to assign high similarity scores to different views of the same data instance, also called positive samples, and low scores for the views coming from different instances, i.e. negative samples. During fine-tuning, a simple classification model is trained on top of the features generated by the pre-trained encoder.

In the following sub-sections, we describe the contrastive learning frameworks, namely unimodal CSSHAR \cite{Khaertdinov_2021_csshar} (Figure \ref{fig:csshar}) and multimodal CMC \cite{Tian_2020_cmc} (Figure \ref{fig:cmc}), and how they can be adapted to Human Activity Recognition.

\subsubsection{CSSHAR: Unimodal Contrastive Learning for Inertial Modality}

\begin{figure*}[!t]
\centering
\scalebox{0.9}{
\includegraphics[width=15cm]{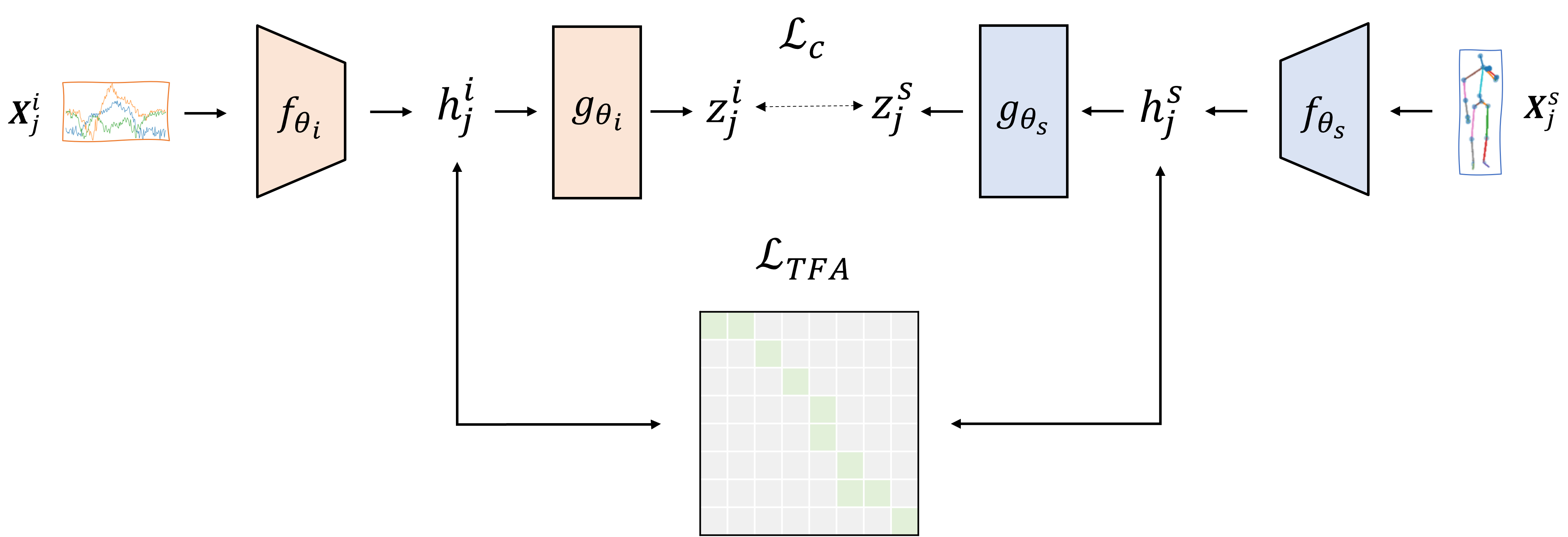}
}
\caption{CMC with Temporal Feature Alignment in the multimodal settings. In the unimodal scenario, the approach two augmented views of the same instance are used as inputs for each stream of the framework.}
\label{fig:cmc-tfa}
\vspace{-15pt}
\end{figure*}

In the unimodal SimCLR framework \cite{Chen_2020_simclr}, or its adaptation to HAR \cite{Khaertdinov_2021_csshar}, different views of the same instance are obtained by applying two random augmentations $t$ and $t'$ to each input signal $\textbf{X}_l^i$ in a mini-batch. Thus, the obtained views $t(\textbf{X}_l^i)$ and $t'(\textbf{X}_l^i)$ form a positive pair. Then, they are fed to an encoder $f_{\theta_i}$ and a projection head $g_{\theta_i}$ to obtain feature representations $z_i^i = g_{\theta_i}(f_{\theta_i}(t(\textbf{X}_l^i)))$ and $z_j^i = g_{\theta_i}(f_{\theta_i}(t'(\textbf{X}_l^i)))$. Here, $z_i^i$ and $z_j^i$ is a positive pair of view-level features for an input inertial signal. Meanwhile, $z_i^i$ and $z_j^i$ form negative pairs with features $z_k^i$ obtained from different instances in the batch. Finally, a matrix with pairwise cosine similarities of the features computed for the whole batch is used to calculate loss. In particular, normalized temperature-scaled cross entropy (NT-Xent) loss aims to contrast positive pair similarity in comparison to similarities of negative pairs present in a batch. The loss function for the representations of positive views $z_i^{i}$ and $z_j^{i}$ is defined as follows:
\begin{equation}
l(i, j) = -log\frac{\delta({z}_i^{i},{z}_j^{i})}{\delta({z}_i^{i},{z}_j^{i})) + \sum_{k=1}^{2n} \mathbb{I}_{[k \neq i, j]} \delta({z}_i^{i},{z}_k^{i})}
\label{eq:ntxent}
\end{equation}
where $\delta({z}_i^{i},{z}_j^{i}) = exp(s({z}_i^{i},{z}_j^{i}))/\tau$, $s(\cdot)$ is a cosine similarity function between $z_i$ and $z_j$, $\tau$ is a temperature parameter, and $n$ is a batch size \cite{Chen_2020_simclr}.

The unimodal contrastive loss function for the whole batch can be written as follows:
\begin{equation}
    \mathcal{L}_{c} = \frac{1}{2N}\sum_{k = 1}^N{(l(z_{2k}^i, z_{2k - 1}^i) + l(z_{2k - 1}^i, z_{2k}^i))}.
    \label{eq:final_loss}
\end{equation}
In equation \ref{eq:final_loss}, it is expected that positive representations have neighboring indices $2k-1$ and $2k$.

\subsubsection{CMC for Inertial and Skeleton Data}

In the multimodal case, each sample $\{\boldsymbol{X}_j^{i},\boldsymbol{X}_j^{s}\}$ in a training batch of size $N$ is passed through modality-specific encoders $f_{\theta_i}, f_{\theta_s}$ and projection heads $g_{\theta_i}, g_{\theta_s}$ to generate projections $\boldsymbol{z}_j^{i}=g_{\theta_i}(f_{\theta_i}(\tilde{\boldsymbol{X}}^i_j))$ and $\boldsymbol{z}_j^{s}=g_{\theta_s}(f_{\theta_s}(\tilde{\boldsymbol{X}}^s_j))$. This pair of features is considered positive. In contrast, all inter-modal combinations of projections from different input instances are treated as negative pairs. Therefore, the loss for an inertial input $\boldsymbol{X}_j^{i}$ and the corresponding representation $\boldsymbol{z}_j^{i}$ is defined as:


\begin{equation}
    l_j^{i \rightarrow s} = - log \frac{\delta({z}_j^{i},{z}_j^{s})}{\sum_{k=1}^N \delta({z}_j^{i},{z}_k^{s})},
    \label{eq:info-nce}
\end{equation}

where, similarly to equation \ref{eq:ntxent}, $\delta({z}_j^{i},{z}_j^{s}) = exp(s({z}_j^{i},{z}_j^{s}))/\tau$. The loss for the $j$-th skeleton representation $l_j^{s \rightarrow i}$ is defined in a similar way.

The total contrastive loss accumulated over the training batch is calculated as follows:

\begin{equation}
    \mathcal{L}_{c} = \frac{1}{2N}\sum_{j=1}^N (l_j^{i \rightarrow s} + l_j^{s \rightarrow i})
    \label{eq:cmc_total_loss}
\end{equation}

\subsection{Temporal Feature Alignment}
In this paper, we introduce a novel temporal feature alignment module that can be integrated into contrastive learning frameworks. Namely, we propose exploiting the Dynamic Time Warping (DTW) algorithm in a feature space to force the alignment of views corresponding to the same instance along the temporal dimension. In Figure \ref{fig:cmc-tfa}, we illustrate the proposed multimodal CMC-TFA algorithm. The unimodal pre-training with CSSHAR-TFA is done in the same manner -- the only difference is that in the unimodal case we use the same encoder for two augmented views of the inertial data input.

Dynamic Time Warping \cite{sakoe_1978_dtw} is an algorithm widely used to compute similarities between time-series data instances \cite{Berndt1994UsingDT}. Instead of calculating the average distance between corresponding timesteps of two sequences, DTW computes the minimum alignment cost between the sequences and it is approached using dynamic programming. Recently, Haresh et al. \cite{Haresh2021LearningBA} used a differentiable version of DTW discrepancy, so called Soft-DTW \cite{cuturi2017softdtw}, as the main loss for self-supervised video representation learning. Namely, they used Soft-DTW to align feature vectors corresponding to separate frames of different videos in a latent space and introduced additional temporal regularization to avoid trivial collapsing solutions. This regularization forces to pull together features of temporally close frames and push away embeddings for temporally far frames. 

Inspired by this idea, we propose to use temporal feature alignment for the problem of unimodal and multimodal HAR by integrating it within a general contrastive learning framework. Hence, unlike in \cite{Haresh2021LearningBA}, we use the Soft-DTW loss without additional regularization for features generated by the main encoders. In our case, trivial solutions are avoided by negative paires used in the NT-Xent loss calculated for the projected features. Specifically, first, representations for inertial signals $h_j^i = f_{\theta_i}({\boldsymbol{X}}^i_j) \in \mathbb{R}^{T_i \times H} = $ and skeletons $h_j^s = f_{\theta_i}({\boldsymbol{X}}^s_j) \in \mathbb{R}^{T_s \times H}$, where $T_i$ and $T_s$ are temporal dimensions of feature representations and $H$ is the number of feature channels, are $l_2$-normalized. Then, the Soft-DTW alignment cost for this pair of embeddings can be defined as follows:

\begin{equation}
    dtw^{\gamma}(h_j^i, h_j^s) = min^{\gamma}_{A \in A_{T_i, T_s}}(A, D),
\label{eq:soft_dtw}
\end{equation}

where $D \in \mathbb{R}^{T_i \times T_s}$ is the distance matrix between feature representations along the temporal dimensions, and $A_{T_i, T_s} \in \{0, 1\}^{T_i \times T_s}$ is the set of all possible alignment matrices. In Equation \ref{eq:soft_dtw}, $min^\gamma$ refers to a soft-min differentiable function \cite{cuturi2017softdtw}. The temporal feature alignment loss for the whole batch can be formulated as follows:

\begin{equation}
    \mathcal{L}_{TFA} = \frac{\sum_{j=1}^N dtw^{\gamma}(h_j^i, h_j^s)}{N}
    \label{eq:tfa}
\end{equation}

The alignment cost for the unimodal settings is defined similarly for each positive pair of features $h_i^i = f_{\theta_i}(t(\textbf{X}_l^i))$ and $h_j^i = f_{\theta_i}(t'(\textbf{X}_l^i))$ obtained for the augmented views.

The final loss consists of the contrastive part and temporal feature alignment component, and could be summarized as follows:

\begin{equation}
    \mathcal{L} = \mathcal{L}_{c} + \alpha \mathcal{L}_{TFA},
    \label{eq:final}
\end{equation}
where $\alpha$ is a TFA weight hyperparameter.

\subsection{Fine-tuning}
The fine-tuning routines for unimodal CSSHAR and multimodal CMC are shown in the right parts of Figures \ref{fig:csshar} and \ref{fig:cmc}, respectively. In the unimodal sensor-based problem, features extracted by the encoder are flattened and directly passed to a classifier, either a simple MLP or a linear classifier. In contrast, for the multimodal case, inertial and skeleton representations are mapped to the same size using a single fusion linear layer, including batch normalization and ReLU. The outputs are then concatenated and passed through the classification linear model.

%% file: sections/4_implementation_details.tex
\section{Experimental setup}
\subsection{Datasets}
\label{sec:datasets}
In this paper, we used 3 widely used unimodal sensor-based HAR datasets as well as one multimodal dataset where we exploited inertial and skeleton modalities. For unimodal datasets pre-processing, we downsample signals to 30Hz frequency, extract 1 second intervals with 50\% overlapping, and normalize signals to zero mean and unit variance per channel. In the multimodal case, the signals are resampled into time-windows of 50 frames. For the skeleton modality, we normalise joint positions in all skeleton sequences based
on the first frame of each sequence.

\noindent\textbf{UCI-HAR} \cite{Anguita_2013_uci} has been acquired using smartphones equipped with IMUs placed on waists of subjects. Overall, 30 subjects took part in data collection performing basic activities of daily living and transitions between them. In this work, as in previous studies, we only focus on 6 main activities. For the holdout test set experiments, we used 20\% of subjects for test and 20\% of them for validation.  

\noindent\textbf{MobiAct} \cite{vavoulas_2016_mobiact} dataset contains data collected from 61 subjects using smartphone placed in their pockets. Overall, 11 daily activities are present in the dataset. As for the UCI-HAR dataset, we used 20\% of subjects for the test set and 20\% of remaining subjects for validation. 

\noindent\textbf{USC-HAD} \cite{zhang_2012_usc} has been collected by 14 users performing 12 activities. The device used for data acquisition is MotionNode IMU placed on hip of each subject. Similar to related works, subjects 11 and 12 were used for validation and subjects 13 and 14 -- for tests.

\noindent\textbf{MMAct} \cite{mmact} is a multimodal dataset that contains 36 activities performed by 20 subjects. In this paper, we use the challenge version\footnote{challenge dataset: \url{https://mmact19.github.io/challenge/} } of the dataset with both inertial signals and 2D-skeletal joints. The dataset provides two evaluation settings, namely cross-subject and cross-scene. The cross-subject scenario is similar to the unimodal datasets holdout sets and 16 first subjects are used for training and validation, while others -- for testing. In the cross-scene setting, all samples collected in the occlusion scene are used for testing, and all other scenes -- for training and validation.

\subsection{Implementations details}
\noindent\textbf{Backbone models.}
In the unimodal sensor-based case, we use 3 layers of 1D-CNN followed by transformer self-attention layers with hyperparameters proposed in \cite{Khaertdinov_2021_csshar}. In the multimodal case, for inertial signals we use the same architecture except that 1D-CNN contains 5 layers. In case of skeleton modality, we adapted a co-occurence feature learning model from \cite{li_2018_cooccurrence} to 2D key-points provided in MMAct. It is also important to mention that we do not apply pooling along the temporal dimension when the proposed Temporal Feature Alignment is used. Supervised models used as baselines in evaluations follow the same encoder architectures as the SSL models. Specifically, unimodal and multimodal supervised models are trained similarly as shown in fine-tuning diagrams in Figures \ref{fig:csshar} and \ref{fig:cmc}, respectively. The only difference is that the encoders are not frozen and trained using backpropagation.

\noindent\textbf{Unimodal pre-training.}
All unimodal encoders are pre-trained within CSSHAR \cite{Khaertdinov_2021_csshar} and the proposed CSSHAR-TFA approach for 200 epochs using LARS optimizer \cite{you_2017_lars} with the batch size 256 and a fixed temperature value of $0.1$. We fix $\gamma=0.1$ from equation \ref{eq:soft_dtw} and $\alpha$ from equation \ref{eq:final} for UCI-HAR, MobiAct and USC-HAD equal to $0.1$, $0.05$ and $0.1$, respectively. Besides, we slightly alter the optimal set of augmentations identified in \cite{Khaertdinov_2021_csshar} and add shift augmentation (by 5-10 timesteps) for MobiAct and use shift instead of permutation for UCI-HAR and USC-HAD in order not to shuffle signals in time multiple times. We also note that the same augmentations have been used while pre-training the CSSHAR approach in order to evaluate the effect of TFA. 

\noindent\textbf{Multimodal pre-training.}
In case of the multimodal CMC \cite{Tian_2020_cmc} and the suggested CMC-TFA approach, pre-training is done in 300 epochs using Adam optimizer \cite{Kingma_2015_Adam}, while the temperature is fixed to $0.1$, $\gamma=0.1$ and $\alpha = 0.005$. We also used modality specific augmentations applied to each instance with fixed probability of $0.75$ in order to artificially enlarge the diversity of the input samples used for pre-training. For the inertial modality, these augmentations are scaling and rotation, while for the skeleton modality we used jittering, resized crops, scaling, rotation and shearing.

\noindent\textbf{Fine-tuning.}
Fine-tuning in both the unimodal and multimodal cases is done in 100 epochs. In order to align our fine-tuning for recent unimodal models we used 3 layer MLP as the classification model with 256 and 128 hidden neurons. Each hidden layer of the classification model also consists of ReLU activation and dropout ($p = 0.2$). Besides that, we also provide results for linear classification when output features are directly mapped to the softmax output layer. In case of multimodal scenario, we use one layer with batch normalization and ReLU activation to map modality-specific features to the same size of 128, concatenate them and then use a linear classifier.

%% file: sections/5_evaluations.tex
\section{Evaluations}
\subsection{Feature Representation Learning}

The first scenario employed in this study is feature representation learning. In order to assess the features learnt during pre-training, we fine-tune a simple classification model on the same training set using features extracted by the frozen pre-trained encoder. In Table \ref{tab:uni_holdout}, we present the macro F1-scores obtained by a simple MLP model on hold-out test sets described in Section \ref{sec:datasets}. In addition, we also perform linear evaluation to compare the proposed model with the adaptation of SimCLR, also known as CSSHAR \cite{Khaertdinov_2021_csshar}. According to the results, the proposed method improves classification performance compared to the previous works on all three datasets using both MLP and linear classifier. 

\input{tables/uni_holdout}

Additionally to the experiments on hold-out test sets, given the problem of subject heterogeneity frequently mentioned in related works \cite{Chen_2021_survey}, we perform 5-fold cross-subject cross-validation. Therefore, each fold was used as a test set, while the remaining folds were used for training. We report the average F1-scores on test folds in Table \ref{tab:uni_cscv}. As can be seen from the table, the TFA module shows better performance than Contrastive Predictive Coding from \cite{haresamudram_2021_cpc_har} and the plain CSSHAR. 

\input{tables/uni_cscv}

The same scenario was employed in the multimodal case on the MMAct dataset using cross-subject and cross-scene scenarios. In Table \ref{tab:multi_holdout}, we compare the classification performance of the proposed approach with the same encoders, pre-trained with CMC and trained in a supervised manner. We also noticed that using augmentations in order to extend the amount of data available for pre-training has a positive effect on feature extraction for both CSSHAR and CSSHAR-TFA. The proposed approach helps to narrow the gap between SSL and supervised models to about 1-2\% F1-score depending on the evaluation scenario employed.


\input{tables/multi_holdout}

\subsection{Semi-supervised learning}

\begin{figure*}[!t]
\centering
\begin{minipage}{.3\linewidth}
\centering
\subfloat[UCI-HAR]{\label{fig:uci_semisup}\includegraphics[scale=.3]{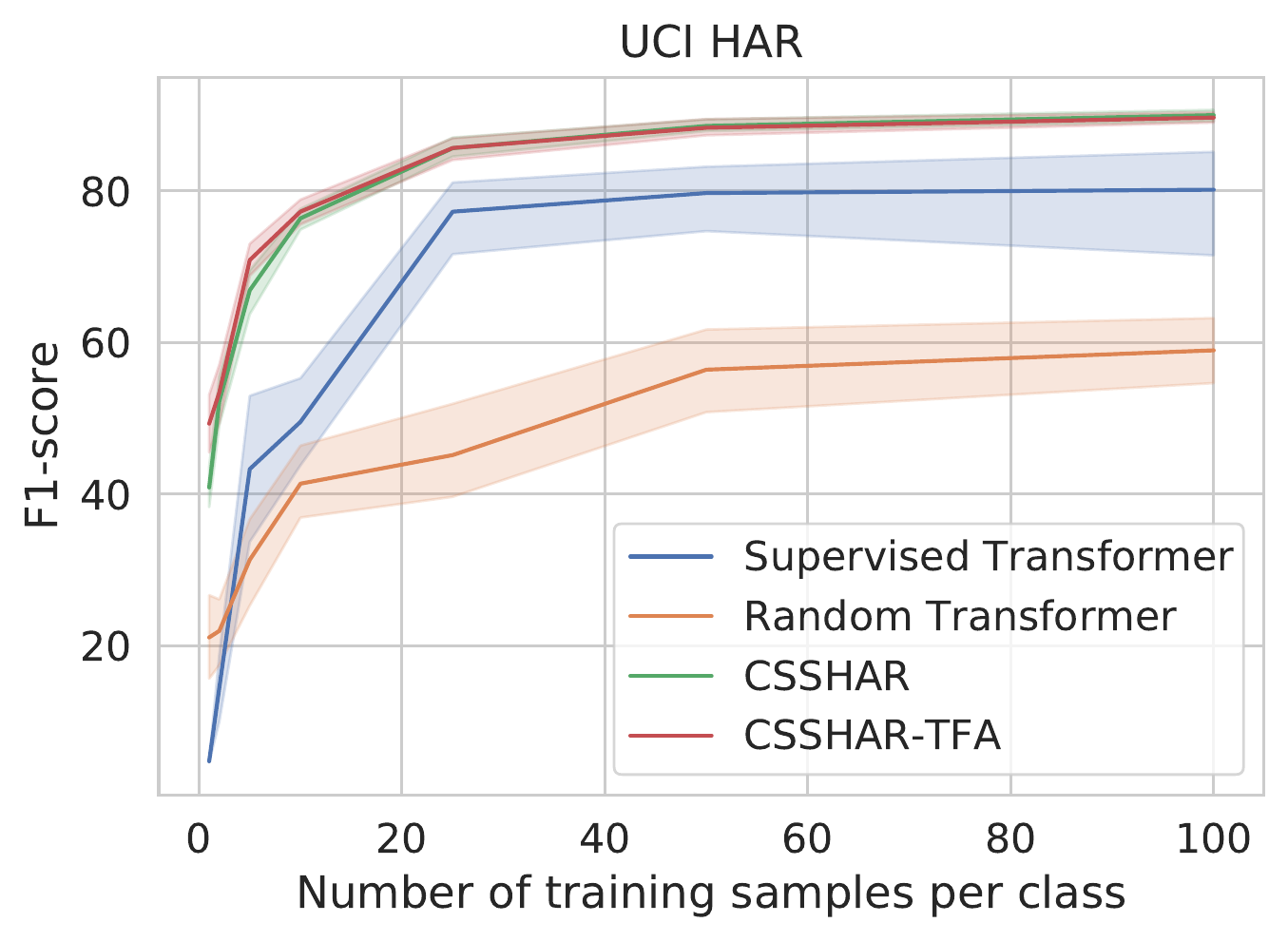}}
\end{minipage}%
\begin{minipage}{.3\linewidth}
\centering
\subfloat[MobiAct]{\label{fig:mobiact_semisup}\includegraphics[scale=.3]{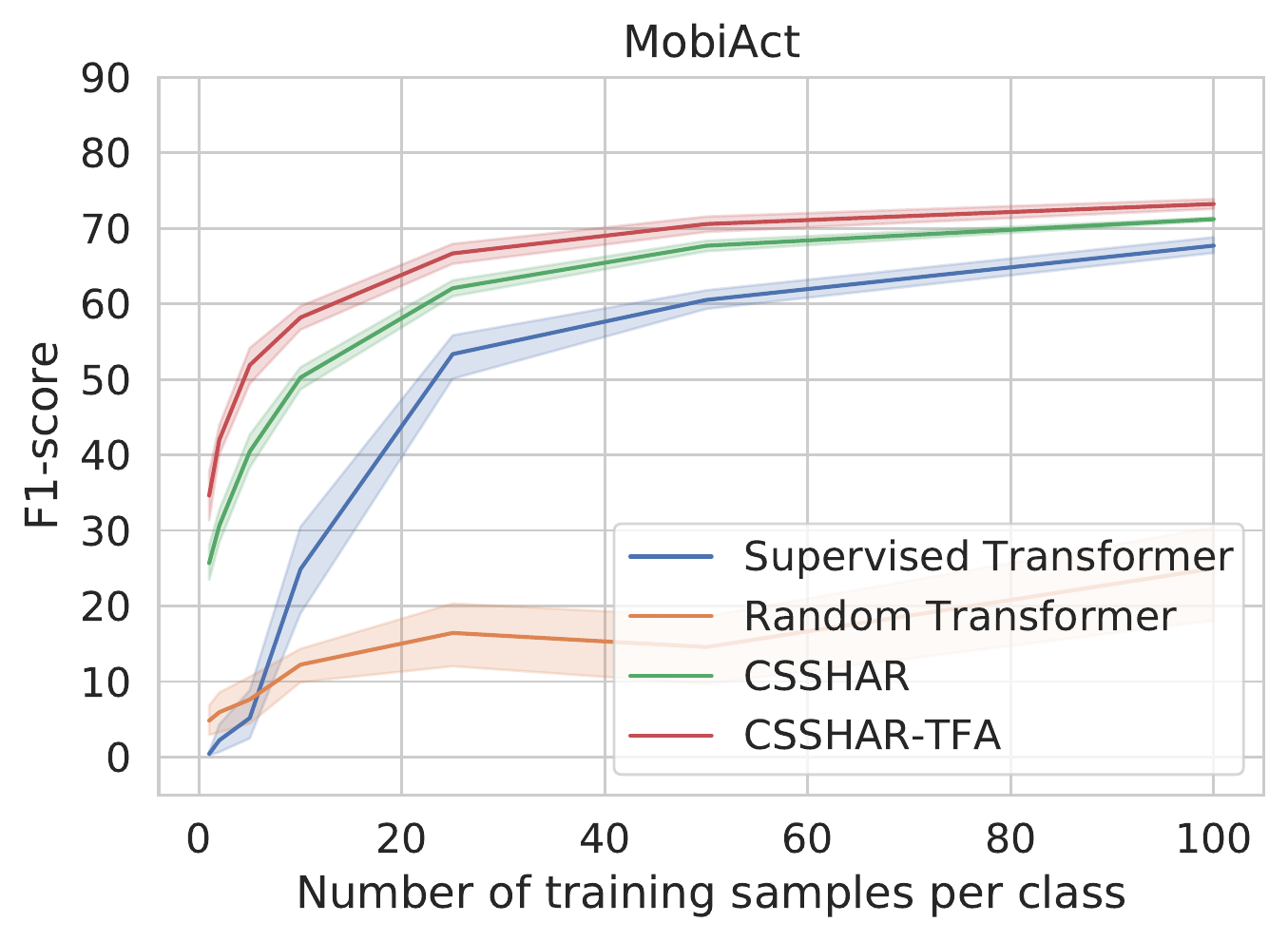}}
\end{minipage}
\begin{minipage}{.3\linewidth}
\centering
\subfloat[USC-HAD]{\label{fig:usc_semisup}\includegraphics[scale=.3]{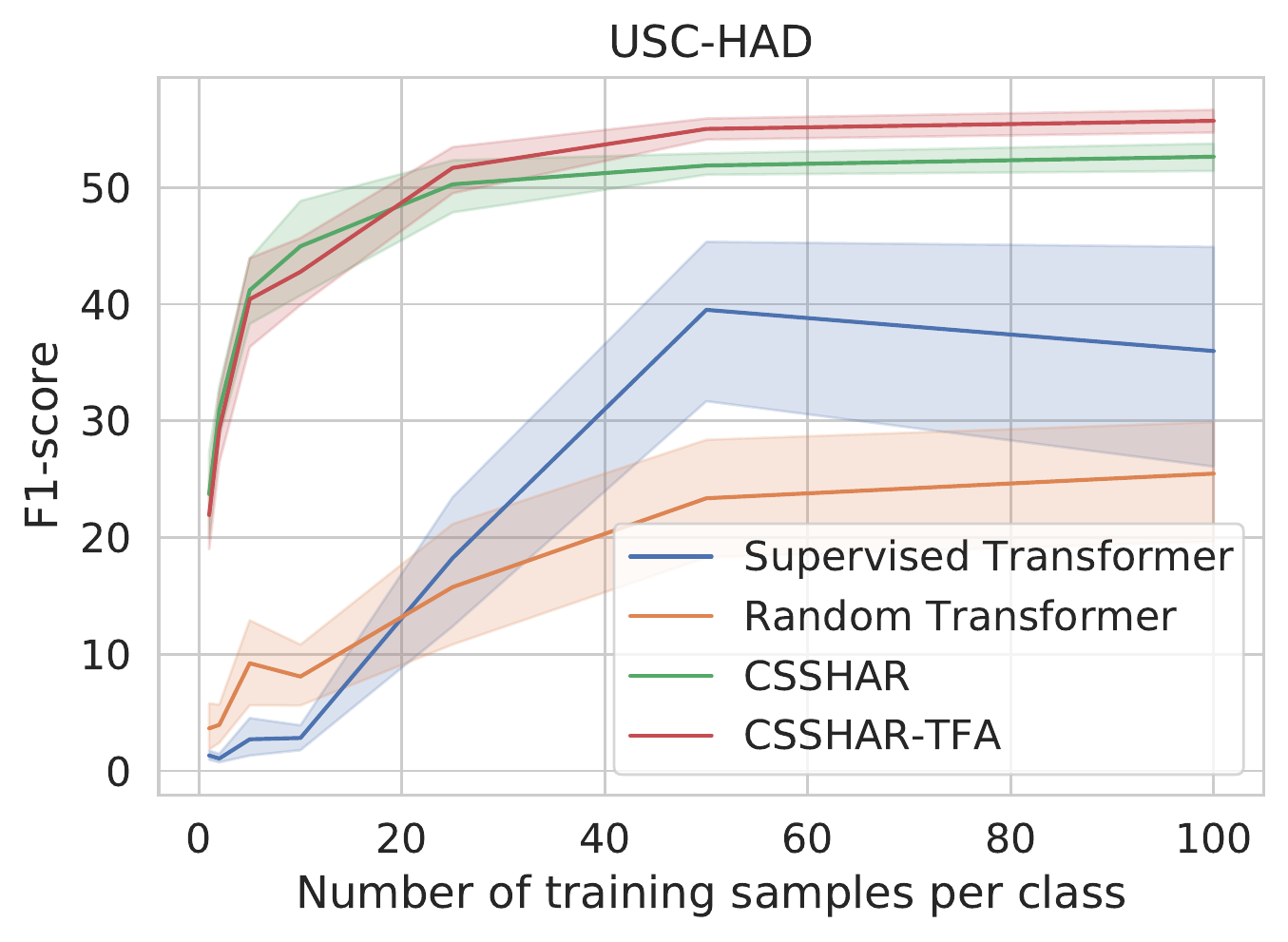}}
\end{minipage}
\begin{minipage}{.3\linewidth}
\centering
\subfloat[MMAct (x-subject)]{\label{fig:mmact_xsub_semisup}\includegraphics[scale=.3]{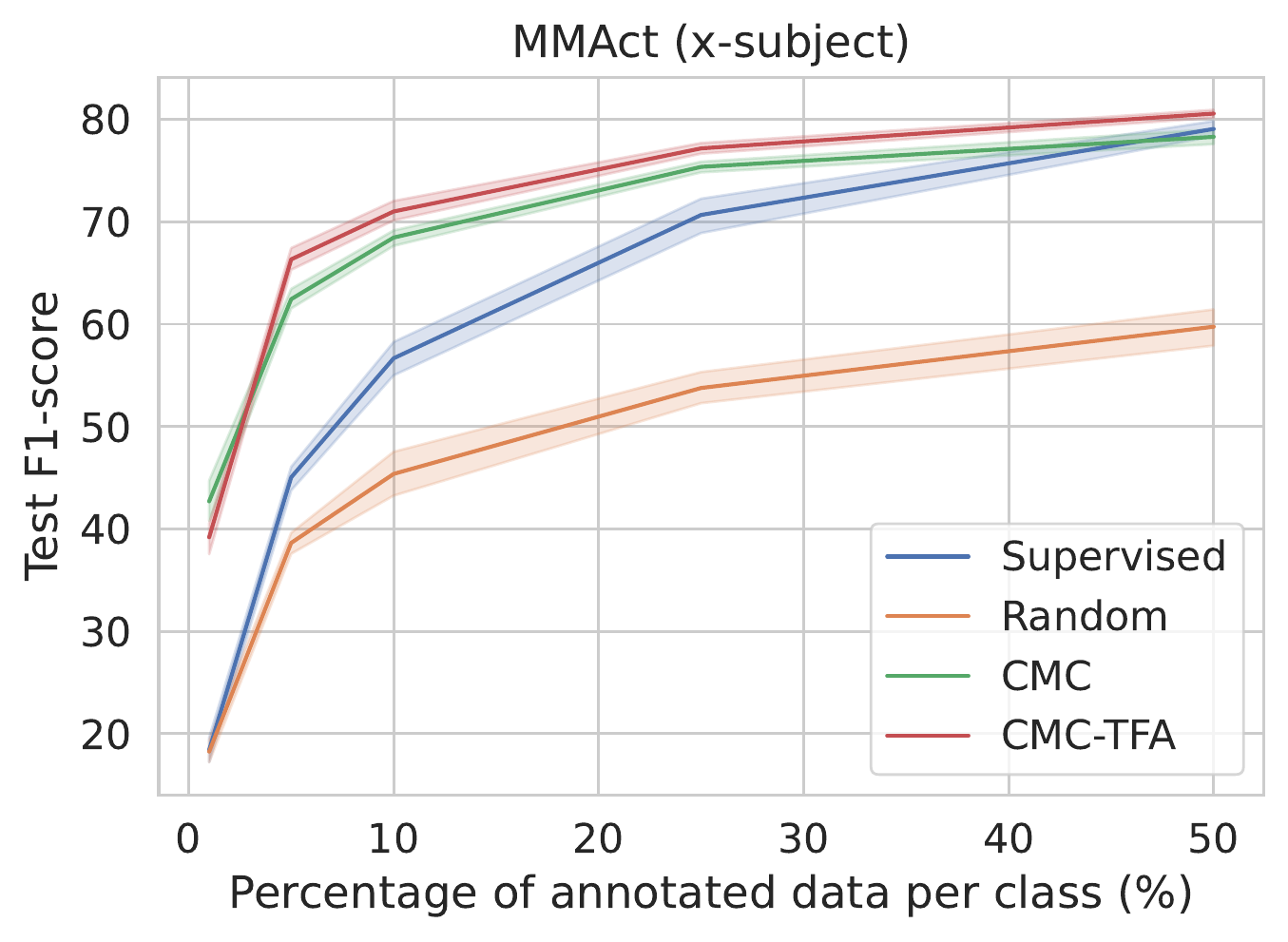}}
\end{minipage}
\begin{minipage}{.3\linewidth}
\centering
\subfloat[MMAct (x-scene)]{\label{fig:mmact_xscene_semisup}\includegraphics[scale=.3]{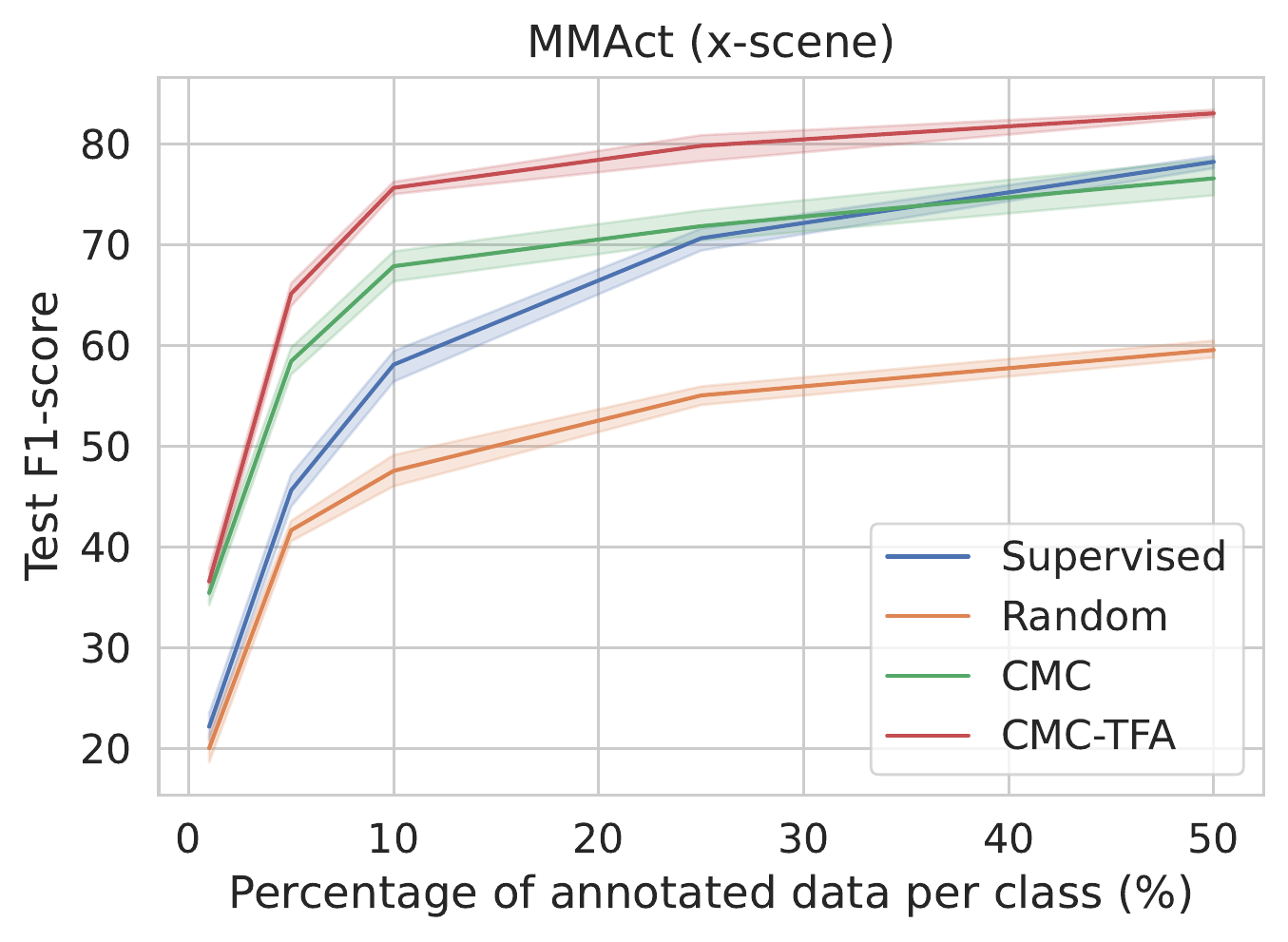}}
\end{minipage}
\caption{Average F1-scores with 95\% confidence intervals for the semi-supervised learning scenario.}
\label{fig:semi}
\end{figure*}

Another scenario used in this study is semi-supervised learning. The main idea is to keep the number of annotated samples limited, for the fine-tuning task. Specifically, a random number or proportion of annotations is selected for fine-tuning. In the unimodal case, we select $k$ annotated examples and use them for fine-tuning and testing. We repeat this experiment ten times for $k \in \{1, 2, 5, 10, 25, 50, 100\}$. Finally, we present the obtained average F1-scores along with the 95\% confidence intervals for the CSSHAR, CSSHAR-TFA, supervised and random models in Figures \ref{fig:uci_semisup}, \ref{fig:mobiact_semisup} and \ref{fig:usc_semisup}. We can see that the TFA module clearly improves performance on MobiAct for almost all values of $k$ and increase on USC-HAD when $k>25$. For the UCI-HAR dataset, CSSHAR and CSSHAR-TFA show comparable performance and confidence intervals intersect for almost all values of $k$.

In the multimodal case, we limit percentage of annotations $p \in \{1\%, 5\%, 10\%, 25\%, 50\% \}$ available for fine-tuning on MMact cross-subject and cross-scene protocols. The obtained results are presented in Figures \ref{fig:mmact_xsub_semisup} and \ref{fig:mmact_xscene_semisup}. Similarly to the unimodal case, the proposed CMC-TFA approach clearly outperforms plain CMC and supervised models in both cross-subject and cross-scene protocols of the MMAct dataset.


\subsection{Discussion on Temporal Alignment}
\begin{figure}[!t]
\centering
\begin{minipage}{.3\linewidth}
\centering
\subfloat[CSSHAR-TFA: self-distance]{\label{fig:tfa_dist_a}\includegraphics[scale=.21]{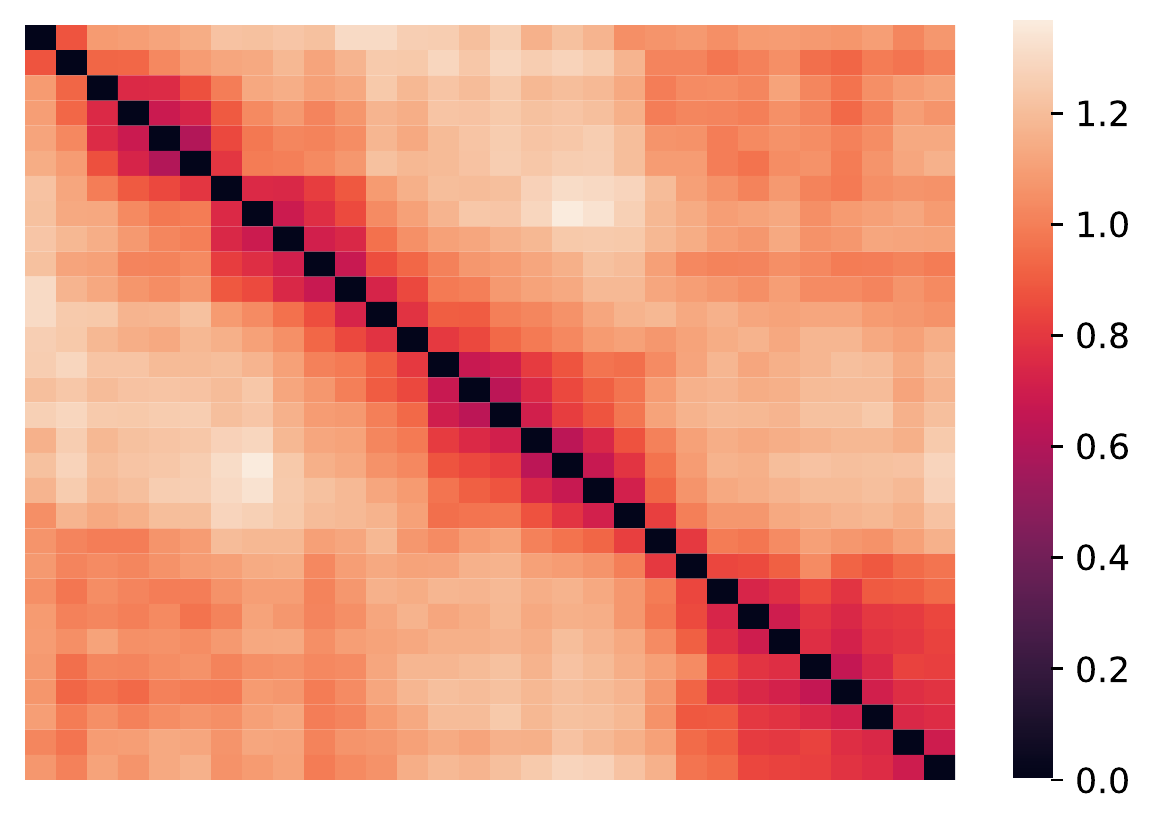}}
\end{minipage}%
\hfill
\begin{minipage}{.3\linewidth}
\centering
\subfloat[CSSHAR-TFA: positive]{\label{fig:tfa_dist_p}\includegraphics[scale=.21]{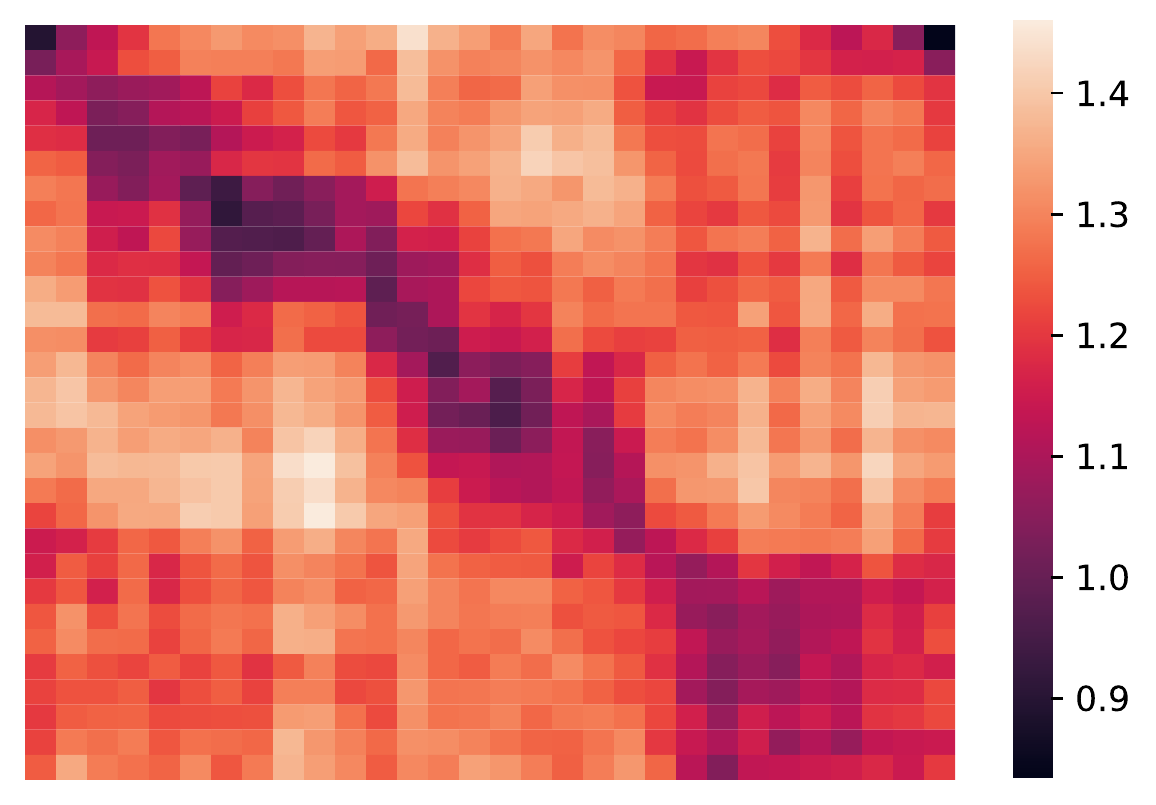}}
\end{minipage}
\hfill
\begin{minipage}{.3\linewidth}
\centering
\subfloat[CSSHAR-TFA: negative]{\label{fig:tfa_dist_n}\includegraphics[scale=.21]{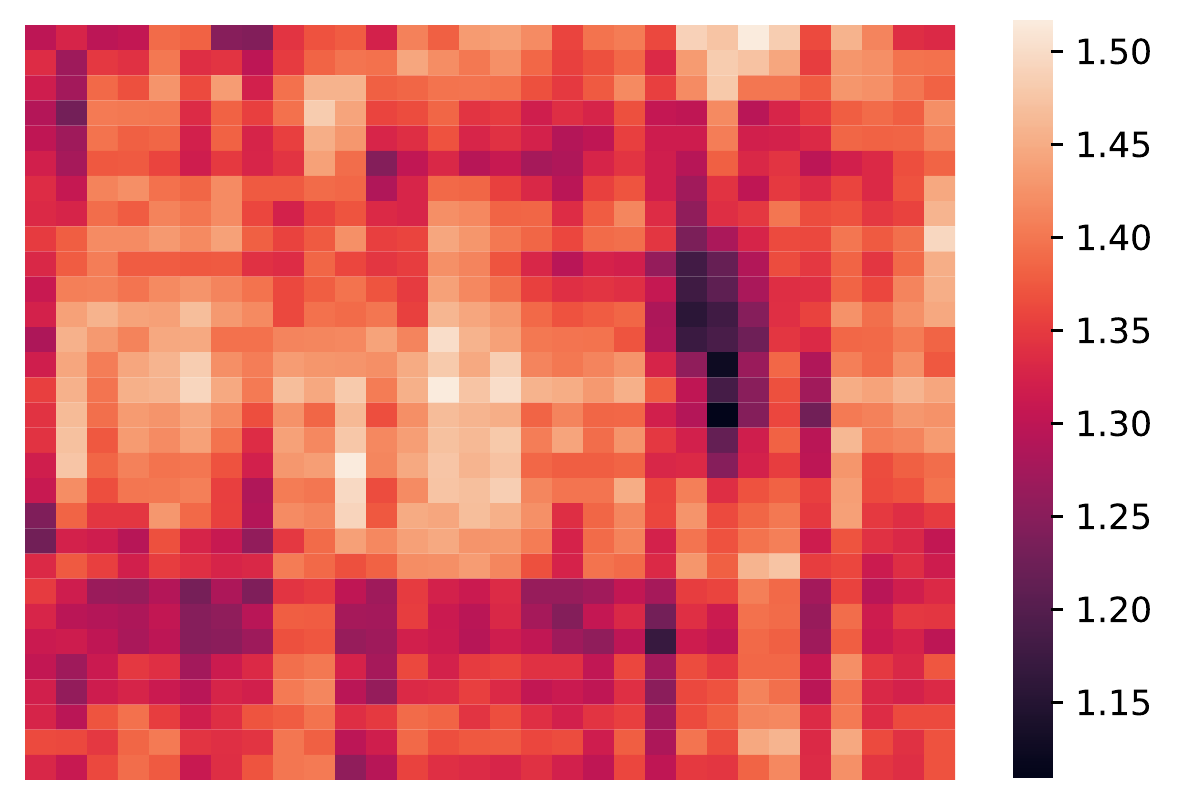}}
\end{minipage}
\begin{minipage}{.3\linewidth}
\centering
\subfloat[CSSHAR: self-distance]{\label{fig:csshar_dist_a}\includegraphics[scale=.21]{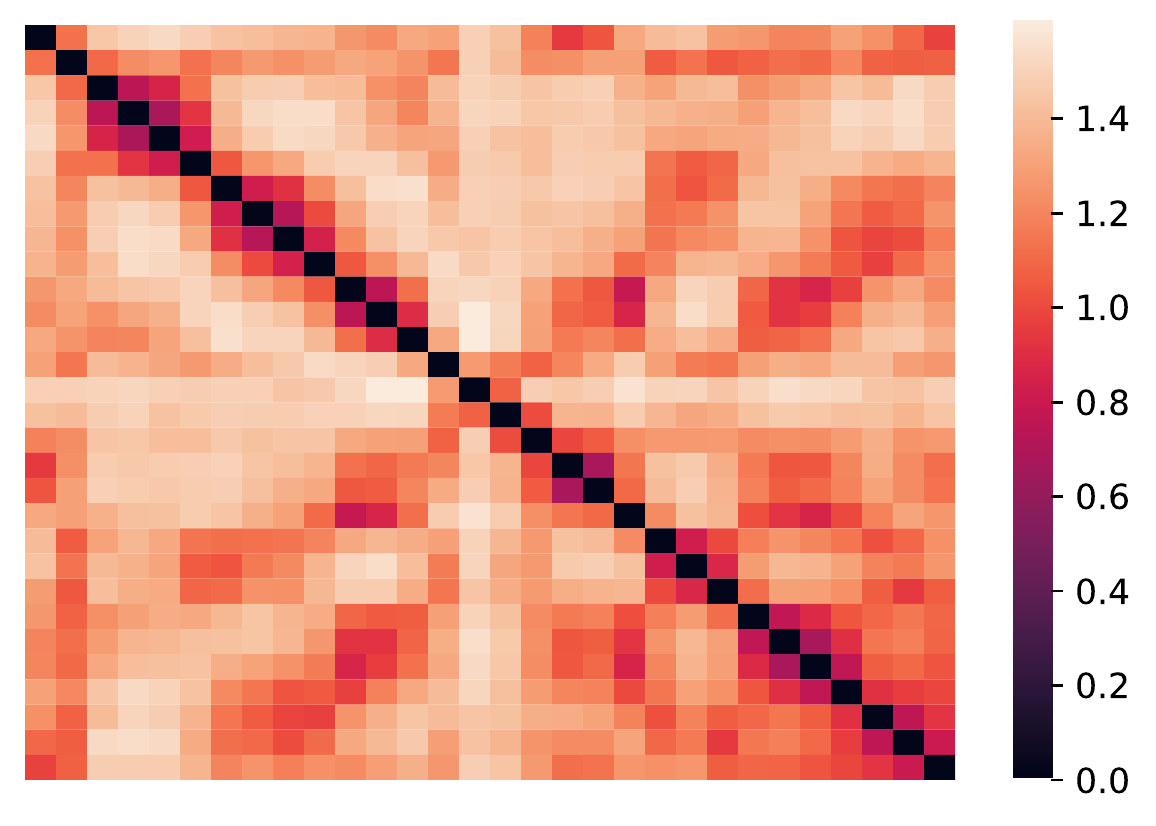}}
\end{minipage}%
\hfill
\begin{minipage}{.3\linewidth}
\centering
\subfloat[CSSHAR: positive]{\label{fig:csshar_dist_p}\includegraphics[scale=.21]{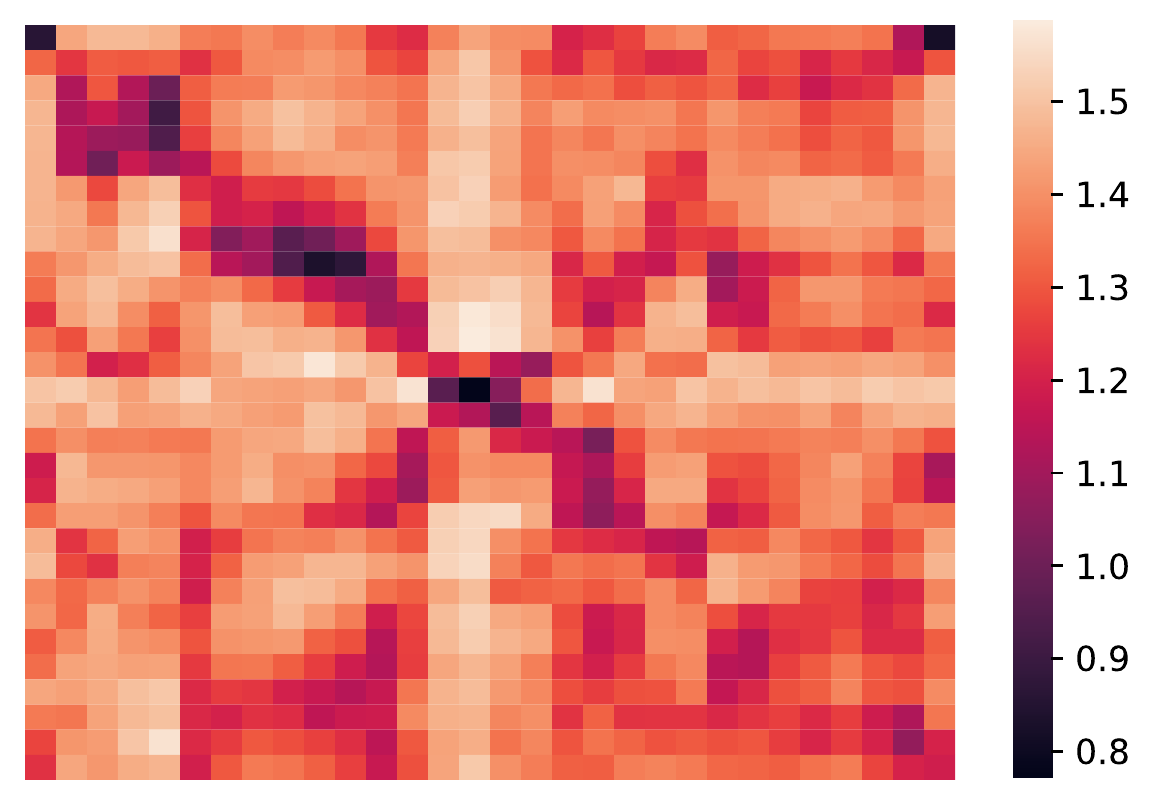}}
\end{minipage}
\hfill
\begin{minipage}{.3\linewidth}
\centering
\subfloat[CSSHAR: negative]{\label{fig:csshar_dist_n}\includegraphics[scale=.21]{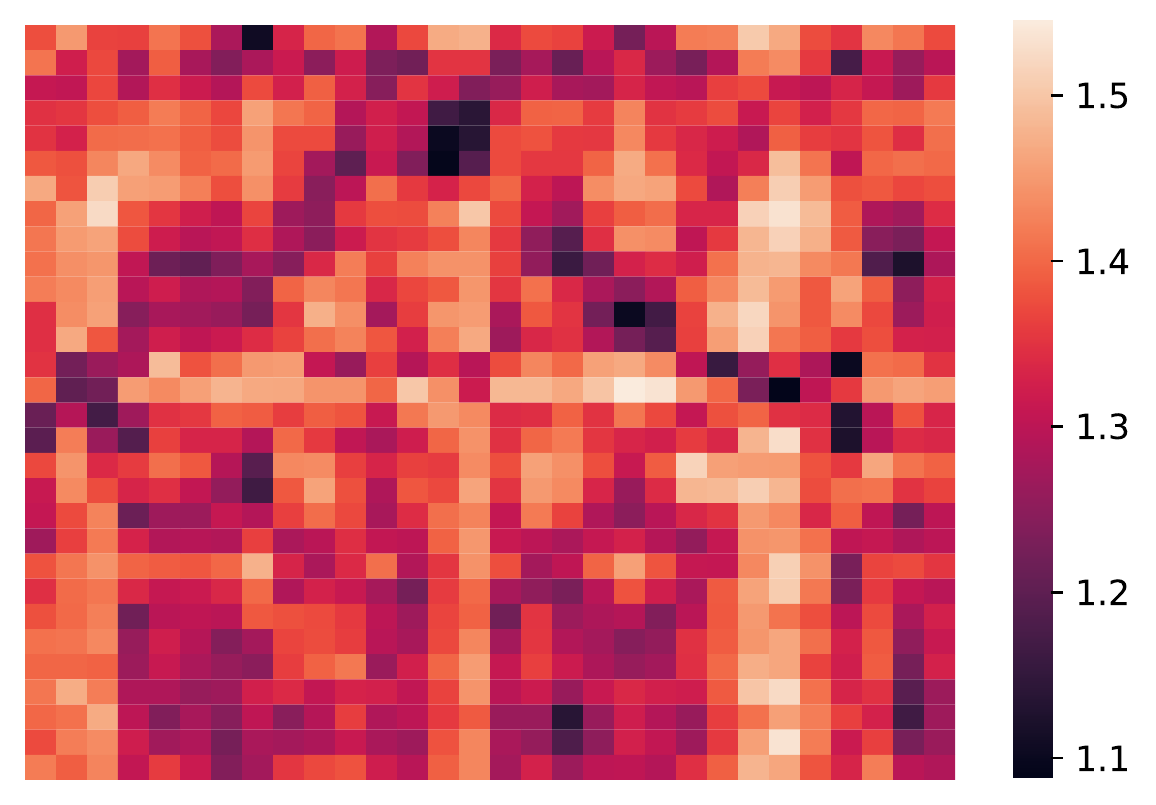}}
\end{minipage}
\caption{Distance matrices between feature embeddings generated by encoders pre-trained using CSSHAR and CSSHAR-TFA for walking (anchor and positive) and jogging (negative) activities from the MobiAct dataset.}
\label{fig:dist_apn}
\end{figure}

Apart from various evaluation scenarios, we also analyze how the proposed TFA module affects temporal alignment of feature representations. For this purpose, as an example, we have picked three time windows from MobiAct with signals from which two correspond to the same walking activity (anchor and positive) and one coming from jogging (negative). Then, we generated embeddings using encoders pre-trained with CSSHAR and CSSHAR-TFA and $l_2$-normalized them for all three examples. Finally, we computed pairwise distances between normalized features along the temporal dimension and visualized them in Figure \ref{fig:dist_apn}. 

As can be seen, in this particular example, the proposed TFA module provides quite clear temporal alignment across the main diagonal for the positive pair. Nevertheless, it is worth mentioning that the plain CSSHAR also tracks some of the temporal dependencies but the distance matrix is more noisy. It is also clear that the negative pairs of representations in Figures \ref{fig:tfa_dist_n} and \ref{fig:csshar_dist_n} are not aligned temporally and, in general, the normalized distances are higher than for the positive pairs. 

\begin{figure}[!h]
\centering
\begin{minipage}{.3\linewidth}
\centering
\subfloat[Jumping]{\label{fig:tfa_dist_bad}\includegraphics[scale=.21]{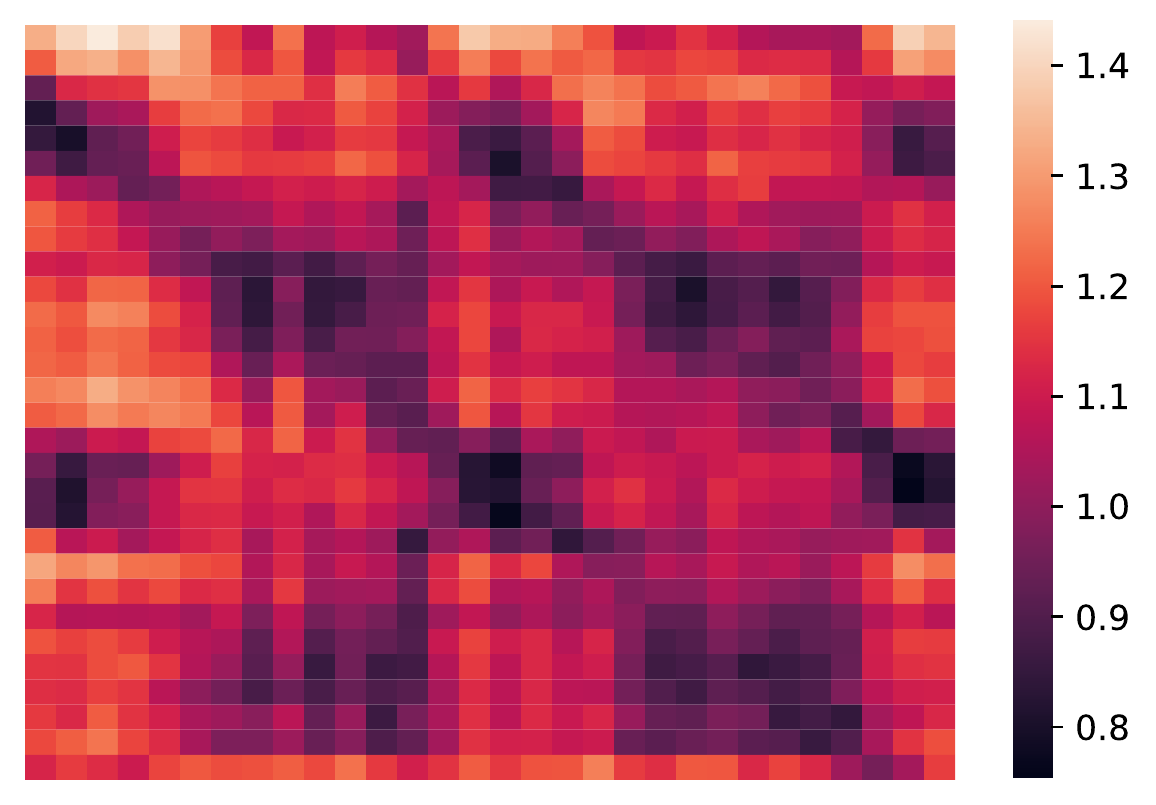}}
\end{minipage}%
\hfill
\begin{minipage}{.3\linewidth}
\centering
\subfloat[Jogging]{\label{fig:tfa_dist_bad1}\includegraphics[scale=.21]{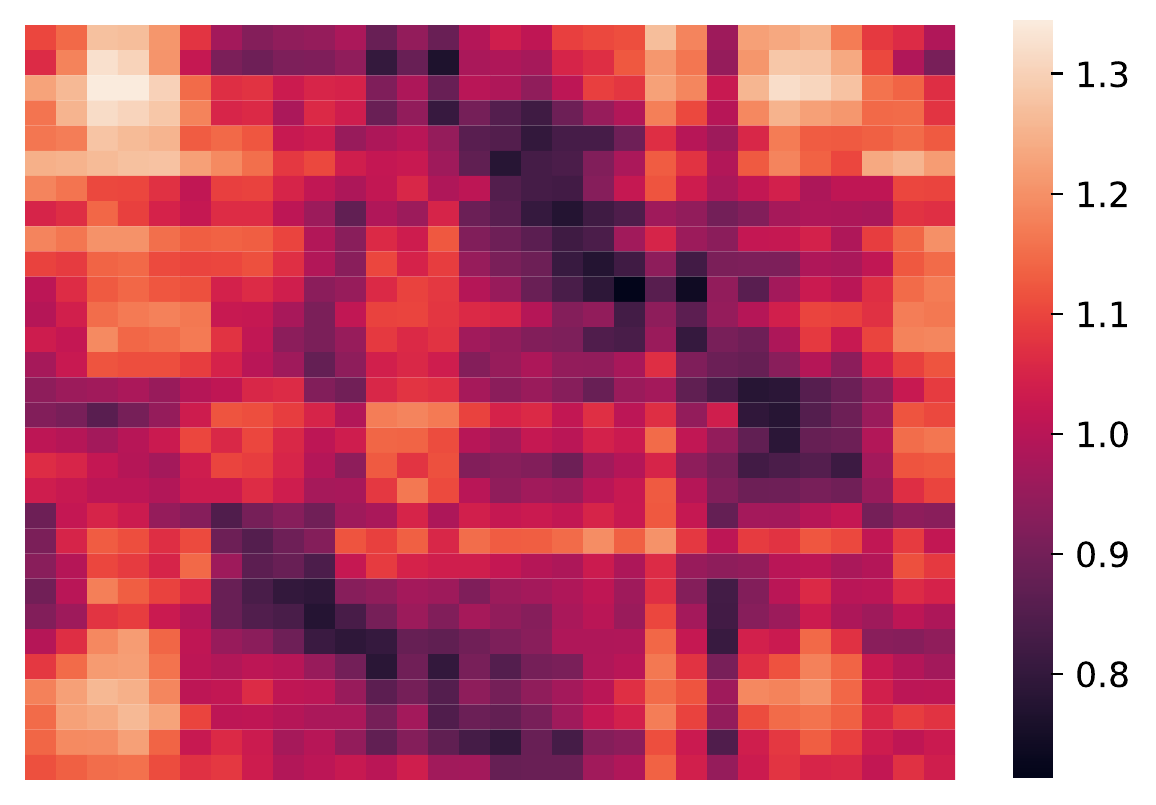}}
\end{minipage}
\hfill
\begin{minipage}{.3\linewidth}
\centering
\subfloat[Going upstairs]{\label{fig:tfa_dist_bad2}\includegraphics[scale=.21]{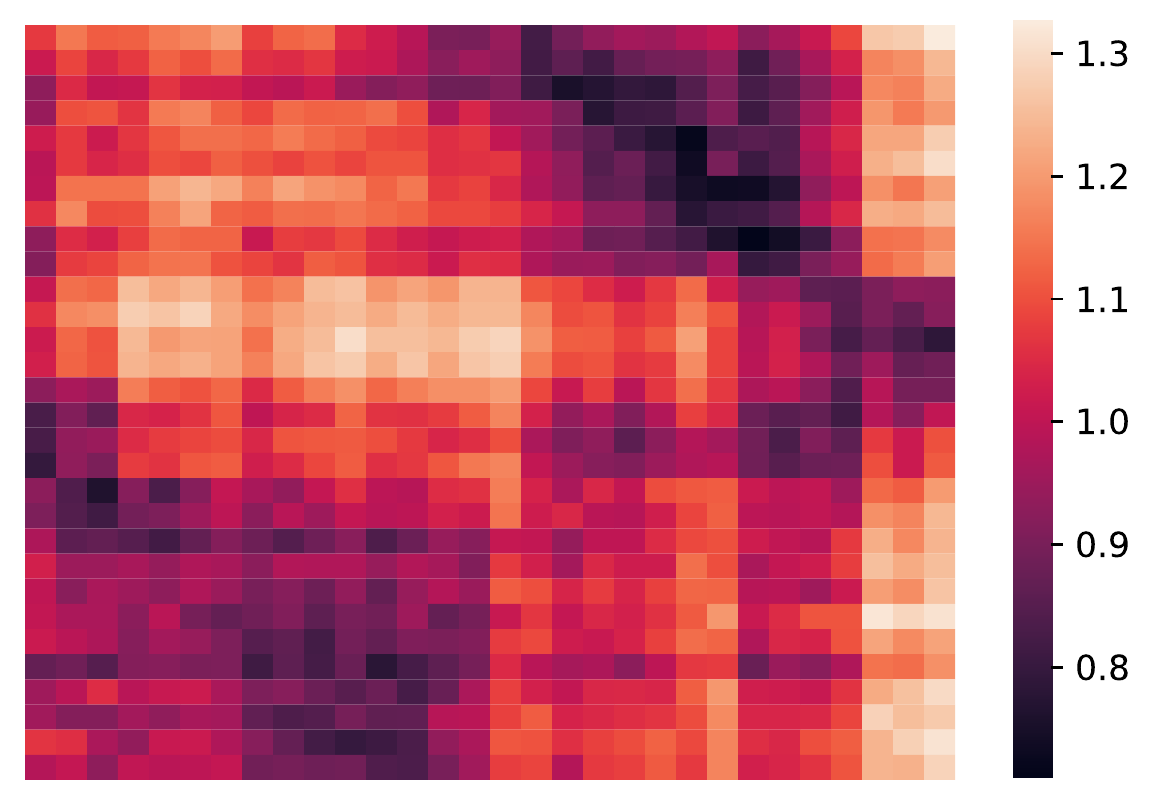}}
\end{minipage}
\caption{Distance matrices for positive pairs of activities with repetitive patterns extracted by the proposed CSSHAR-TFA approach on the MobiAct dataset.}
\label{fig:rep}
\end{figure}

While the proposed algorithm has a positive effect on feature extraction according to various scenarios employed, it does not take into account that some of the activities have repetitive patterns (e.g. jumping, jogging), and it might be the case when these patterns appear multiple times inside the sampled time-windows. A few positive distance matrices for these cases are demonstrated in Figure \ref{fig:rep}. Hence, although the proposed approach still tracks these patterns, it might be a valid idea for the future work to enhance the current approach by identifying these patterns and force temporal alignment within these repeating intervals or sample input instances differently for activities with different frequencies of patterns.

%% file: tables/uni_holdout.tex
\begin{table}[!t]
\centering
\scalebox{0.8}{
\begin{tabular}{l|ccc}
                                                                 & UCI-HAR                & MobiAct                & USC-HAD                \\ \hline
Multi-task \cite{Saeed_2019_multitask}                                                       &   80.2                 &   75.41                & 45.37                  \\
CAE \cite{haresamudram_2019_cae}                                                           & 80.26                  &     79.58              & 48.82                  \\
Masked Rec. \cite{haresamudram_2020_maskedrec}                                                     & 81.89                  &    76.81               & 49.31                  \\
CSSHAR \cite{Khaertdinov_2021_csshar}                                                          & 91.96 (90.59)          & 82.65 (82)           & 56.56 (55.56)          \\
CSSHAR-TFA                                                       & \textbf{93.42 (90.99)} & \textbf{83.76 (83.44)} & \textbf{57.81 (56.28) } \\ \hline
\begin{tabular}[c]{@{}l@{}}Supervised\\ Transformer\end{tabular} & 95.26                  & 83.92                  & 60.56                 
\end{tabular}
}
\caption{Mean F1-scores for the unimodal sensor-based activity recognition task. Values in brackets for CSSHAR and the proposed CSSHAR-TFA models correspond to the linear evaluation scenario.}
\label{tab:uni_holdout}
\end{table}

%% file: tables/uni_cscv.tex
\begin{table}[!h]
\centering
\scalebox{0.9}{
\begin{tabular}{l|ccc}
           & UCI-HAR        & MobiAct        & USC-HAD       \\ \hline
CPC \cite{haresamudram_2021_cpc_har}       & 90.24 & 76.24          & 65.15         \\
CSSHAR \cite{Khaertdinov_2021_csshar}     & 90.02          & 81.16          & 72.39         \\
CSSHAR-TFA & \textbf{90.7}         & \textbf{83.23} & \textbf{72.74}
\end{tabular}}
\caption{Average F1-scores for 5-fold cross-subject cross-validation.}
\label{tab:uni_cscv}
\end{table}

%% file: tables/multi_holdout.tex
\begin{table}[!t]
\centering
\begin{tabular}{l|ccc}
                          & augmentations & x-scene                               & x-subject                             \\ \hline
                          & \ding{55}            & { 80.86}          & { 77.04}          \\
\multirow{-2}{*}{CMC}     & \ding{51}            & { 83.61}          & { 82.02}         \\
                          & \ding{55}             & { 81.44}         & { 79.9}         \\
\multirow{-2}{*}{CMC-TFA} & \ding{51}            & { \textbf{85.14}} & { \textbf{83.36}} \\ \hline
Supervised                & \ding{55}             & { 87.36}          & { 84.05}
\end{tabular}
\caption{Activity recognition F1-scores on the multimodal MMAct dataset.}
\label{tab:multi_holdout}
\end{table}

%% file: sections/6_conclusion.tex
\section{Conclusion}
In this paper, we presented a novel Temporal Feature Alignment module that can be integrated into unimodal and multimodal contrastive learning frameworks for Human Activity Recognition. Specifically, we integrated the proposed approach into SimCLR and CMC. The conducted experiments have shown that the proposed method has a good potential to improve representations learnt during pre-training as has been demonstrated in various evaluation scenarios. We also qualitatively assessed the temporal alignment of features by visualizing the encoded features for the unimodal case. We identified that the proposed method identifies different types of patterns in data. The future work may focus on more sophisticated techniques to explicitly support temporal alignment of signals having these patterns.

%% file: sections/7_acknowledgement.tex
\section*{Acknowledgement}
This work has been funded by the European Union’s Horizon2020 project: PeRsOnalized Integrated CARE Solution for Elderly facing several short or long term conditions and enabling a better quality of LIFE (Procare4Life), under Grant Agreement N.875221.